\documentclass[11pt]{article}

\usepackage[preprint]{acl}

\usepackage{times}
\usepackage{latexsym}
\usepackage{float}

\usepackage[T1]{fontenc}

\usepackage[utf8]{inputenc}

\usepackage{microtype}

\usepackage{inconsolata}
\usepackage{booktabs}
\usepackage[most]{tcolorbox}
\usepackage[dvipsnames]{xcolor} 

\usepackage{graphicx}
\usepackage{amsmath} 
\usepackage{amssymb} 
\usepackage{algorithm} 
\usepackage{algpseudocode} 
\newcommand{\procname}[1]{\textsc{#1}}

\algrenewcommand\algorithmiccomment[1]{\hfill\(\triangleright\) #1}

\title{Automated Refinement of Essay Scoring Rubrics \\for Language Models via Reflect-and-Revise}

\author{
 \textbf{Keno Harada},
 \textbf{Lui Yoshida},
 \textbf{Takeshi Kojima},
 \textbf{Yusuke Iwasawa},
 \textbf{Yutaka Matsuo}
\\
The University of Tokyo
\\
 \small{
 \texttt{{keno.harada@weblab.t.u-tokyo.ac.jp}}}
 }

\begin{document}
\maketitle
\begin{abstract}
The performance of Large Language Models (LLMs) is highly sensitive to the prompts they are given. Drawing inspiration from the field of prompt optimization, this study investigates the potential for enhancing Automated Essay Scoring (AES) by refining the scoring rubrics used by LLMs. Specifically, our approach prompts models to iteratively refine rubrics by reflecting on models' own scoring rationales and observed discrepancies with human scores on sample essays. Experiments on the TOEFL11 and ASAP datasets using GPT-4.1, Gemini-2.5-Pro, and Qwen-3-Next-80B-A3B-Instruct show Quadratic Weighted Kappa (QWK) improvements of up to 0.19 and 0.47, respectively. Notably, even with a simple initial rubric, our approach achieves comparable or better QWK than using detailed human-authored rubrics. Our findings highlight the importance of iterative rubric refinement in LLM-based AES to enhance alignment with human evaluations.
\end{abstract}

\section{Introduction}
Automated Essay Scoring (AES) systems powered by Large Language Models (LLMs) are increasingly expected to provide real-time, scalable feedback for students and alleviate the grading burden on instructors~\citep{mizumoto2023aes,yancey-etal-2023-rating,naismith-etal-2023-automated,pack2024validity}. Typically, these systems employ static, pre-defined rubrics to guide the evaluation. However, it remains an open question whether rubrics designed for human raters are optimal for LLMs. When human raters use a rubric, they often engage in a collaborative calibration process: they score sample essays, discuss discrepancies in their judgments, and refine their shared understanding of the criteria to ensure consistency~\citep{trace2016rubricsnegotiate,ozfidan2022rubricdev,ouyang2022instructgpt,yoo-etal-2025-dress}. This iterative, reflective practice is overlooked in current LLM-based AES, potentially limiting their alignment with human scoring patterns.

\begin{figure}[t]
  \includegraphics[width=\columnwidth]{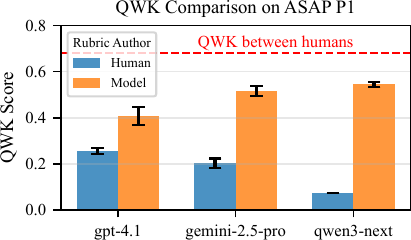}
  \caption{We show the results for a model using a rubric that was automatically refined versus the same model using human-written rubric. The evaluation was performed three times on the test set to ensure reliability, and we report the mean and standard deviation of the QWK scores.}
  \label{fig:experiments}
\end{figure}

Recent studies show that LLMs have abilities to refine their own outputs especially when there are reliable external feedback~\citep{madaan2023selfrefine,kamoi2024selfcorrect}. Prompt optimization techniques leverage these capabilities to update prompts to maximize a targeted metric and show performance improvements in various tasks such as multi-hop reasoning, instruction following and privacy-aware delegation~\citep{khattab2023dspy,opsahl-ong-etal-2024-optimizing,agrawal2025gepa}.

Inspired by these developments and the calibration process of human raters, we propose a iterative refinement approach for rubrics in LLM-based AES.
Specifically, for each model and dataset, using 200 sample essays and scores assigned by human raters, models are prompted to iteratively refine the rubric reflecting on models' own scoring rationales and observed score discrepancies with human scores to mitigate Quadratic Weighted Kappa (QWK) between LLM and human scores. 

Experiments on the TOEFL11~\citep{blanchard2013toefl} and ASAP~\citep{ben2012asap} datasets using GPT-4.1~\citep{gpt4.1}, Gemini-2.5-Pro~\citep{gemini25}, and Qwen-3-Next-80B-A3B-Instruct~\citep{qwen3-next} demonstrate that our approach improves QWK with maximum gains of 0.19 and 0.47 on the respective datasets. Furthermore, we show that even with a simple initial rubric \textit{``Based on the response's content, rate the response on a scale of 1 to 6.''}, our approach achieves comparable or better QWK than using a detailed human-authored rubric, suggesting that LLMs can autonomously identify relevant evaluation criteria. Our findings highlight the importance of iterative refinement in rubric-based LLM AES to enhance alignment with human evaluations.

\section{Related Work}
For non-verifiable tasks, where judging success is not as straightforward as in math or code, recent research has focused on LLM-based automatic evaluation using checklists and rubrics in prompts~\citep{min-etal-2023-factscore,qin-etal-2024-infobench,lin2024wildbench,wu-etal-2025-lifbench,cook2025ticking,huang2025rubricanchors,gunjal2025rubricsasrewards,viswanathan2025checklistsbetterrewardmodels,lee2025checkeval}. AES is an example of such a non-verifiable task, and various techniques have been proposed~\citep{mizumoto2023aes,xie2024gradelikehuman,lee-etal-2024-unleashing}.

Regarding the relationship between rubric design and LLM performance, \citet{furuhashi2025checklist} reported on "negative items"—rubric components that, while valid for human evaluators, do not improve the performance. In fact, they found that removing these items can even improve performance. Furthermore, \citet{yoshida2025rubrics} reported that making rubrics more detailed does not always lead to performance gains in AES.

These findings indicate that a better rubric for an LLM may exist, but its characteristics are not yet understood. While existing research has explored methods such as having humans edit rubrics or having LLMs generate them, these approaches do not typically involve prompting the LLM to reflect on its own output to iteratively improve the rubric for better agreement with human scores. Our method, in contrast, enables the model to perform this iterative refinement process. It updates the rubric based on its scoring performance on sample essays, effectively mimicking the way human evaluators refine their interpretations and build a shared understanding before a formal scoring session~\citep{trace2016rubricsnegotiate,ozfidan2022rubricdev,ouyang2022instructgpt,yoo-etal-2025-dress}. By systematically optimizing the rubric, our work addresses the unexplored potential of tailoring evaluation criteria specifically for LLMs to enhance scoring accuracy and reliability.

\section{Iterative Rubric Refinement}
\label{sec:method}

We propose an iterative refinement algorithm that updates a rubric based on its performance on a held-out validation set. The overall process is detailed in~\autoref{alg:rubric_refinement}.

\subsection{Preliminaries}

Our method requires a dataset, a set of initial seed rubrics, and a language model. The dataset \( \mathcal{D} \) consists of pairs \( (x, y) \), where \( x \) is an essay and \( y \) is its human-assigned holistic score. We split this dataset into three subsets: a training set \( \mathcal{D}_{\text{train}} \), a validation set \( \mathcal{D}_{\text{val}} \), and a test set \( \mathcal{D}_{\text{test}} \). The training set is used to generate feedback for rubric refinement, the validation set is used for selecting the best rubric at each iteration, and the test set is reserved for the final evaluation. 

We prepare a set of seed rubrics, \( \mathcal{R}_{\text{seed}} \), to investigate the impact of the starting point. This set includes: a \texttt{simplest\_rubric}, which only defines the score scale; the \texttt{human\_rubric}, which is the official detailed scoring guide; and a \texttt{simplified\_human\_rubric}, which is a simpler rubric of \texttt{human\_rubric}, similar to~\citet{yoshida2025rubrics}.

The core component is a language model, denoted as \( \mathcal{M} \), tasked with two primary functions. First, in the "scoring" function, \( \mathcal{M} \) takes a rubric and an essay \( x \) as input to generate a predicted score \( \hat{y} \) along with a textual rationale \( z \) explaining its reasoning. Second, in the "refinement" function, \( \mathcal{M} \) is prompted to generate an improved rubric by considering the previous rubric, the generated rationales, and the discrepancies between predicted and ground-truth scores.

\subsection{Iterative Refinement Algorithm}

The refinement process begins by initializing the best rubric, \( R_{\text{best}} \), with chosen seed rubric (\( \mathcal{R}_{\text{seed}} \)). Its initial performance, \( \text{QWK}_{\text{best}} \), is calculated by evaluating it on the validation set \( \mathcal{D}_{\text{val}} \). QWK is a widely-used metric in AES for assessing inter-rater agreement.

The main algorithm then proceeds in a loop for a predefined number of iterations, \( T \). In each iteration, we first sample a mini-batch \( \mathcal{B} \) from the training set \( \mathcal{D}_{\text{train}} \). For each essay \( x_i \) in this batch, we use the current best rubric \( R_{\text{best}} \) to instruct \( \mathcal{M} \) to generate a score \( \hat{y}_i \) and a rationale \( z_i \). 

Next, in the refinement step, we gather all the information from the batch—the rationales, predicted scores, and corresponding ground-truth scores \( y_i \)—and use it to construct a prompt for the LLM. This prompt asks the model to generate a new candidate rubric, \( R_{\text{new}} \), designed to address the observed scoring inaccuracies.

Following this, in the evaluation and selection step, the performance of \( R_{\text{new}} \) is measured by calculating its QWK score on the validation set \( \mathcal{D}_{\text{val}} \). If this new score, \( \text{QWK}_{\text{new}} \), surpasses the current best score, \( \text{QWK}_{\text{best}} \), we update \( R_{\text{best}} \leftarrow R_{\text{new}} \).

After \( T \) iterations, the algorithm returns the final \( R_{\text{best}} \). Our algorithm is a simplified version of~\citet{agrawal2025gepa}, where we have no Pareto-based Candidate Filtering and System Aware Merge for ease of implementation. To analyze the influence of the initial rubric, we report the results of separate optimization runs starting from each of the seed rubrics in \( \mathcal{R}_{\text{seed}} \) in Results section.
\begin{figure}[t]
  \centering
\begin{algorithm}[H]
    \caption{Iterative Rubric Refinement}
    \label{alg:rubric_refinement}
    \begin{algorithmic}[1]
        \Statex \textbf{Require:} Dataset \( \mathcal{D} \), Large Language Model \( \mathcal{M} \), Initial rubric \( R_{\text{seed}} \)
        \Statex \textbf{Require:} Number of iterations \( T \), Batch size \( b \)
        \Statex

        \State \( R_{\text{best}} \leftarrow R_{\text{init}} \) 
        \State \( \text{QWK}_{\text{best}} \leftarrow \procname{Evaluate}(\mathcal{M}, R_{\text{best}}, \mathcal{D}_{\text{val}}) \)
        \Statex

        \For{\(t = 1 \text{ to } T\)}
            \State \( \mathcal{B} \leftarrow \procname{SampleMiniBatch}(\mathcal{D}_{\text{train}}, b) \) 
            \State \( \text{FbData} \leftarrow \emptyset \)
            \For{each \((x, y) \in \mathcal{B}\)}
                \State \( (\hat{y}, z) \leftarrow \procname{Score}(\mathcal{M}, R_{\text{best}}, x) \)
                \State Add \( (\text{rationale}=z, \text{pred\_score}=\hat{y}, \text{true\_score}=y) \) to FbData
            \EndFor
            \Statex

            \State \( R_{\text{new}} \leftarrow \procname{Refine}(\mathcal{M}, R_{\text{best}}, \text{FbData}) \)
            \State \( \text{QWK}_{\text{new}} \leftarrow \procname{Evaluate}(\mathcal{M}, R_{\text{new}}, \mathcal{D}_{\text{val}}) \)
            \Statex

            \If{\( \text{QWK}_{\text{new}} > \text{QWK}_{\text{best}} \)}
                \State \( R_{\text{best}} \leftarrow R_{\text{new}} \)
                \State \( \text{QWK}_{\text{best}} \leftarrow \text{QWK}_{\text{new}} \)
            \EndIf
        \EndFor
        \Statex
        
        \State \textbf{return} \( R_{\text{best}} \)
    \end{algorithmic}
\end{algorithm}
\caption{We propose iterative refinement algorithm that updates a rubric based on its performance on a held-out validation set.}
\end{figure}

\begin{table}[t]
\centering
\begin{tabular}{llll}
\toprule
Refine & Rubric & ASAP & TOEFL \\
\midrule
Yes & from human & \underline{0.46} & 0.56 \\
Yes & from simplified & 0.41 & 0.58 \\
Yes & from simplest & \textbf{0.48} & \textbf{0.64} \\
\hline
No & human & 0.26 & 0.58 \\
No & simplified & 0.33 & \underline{0.59} \\
No & simplest & 0.17 & 0.57 \\
\bottomrule
\end{tabular}
\caption{Even starting from a simplest rubric, “Based on the response’s content, rate the response on a scale of 1 to 6.", the refined rubric can outperform a carefully crafted human-written rubric. }
\label{tab:seed_rubric}
\end{table}

\section{Experiments}
We conducted experiments on two essay scoring datasets: TOEFL11 and ASAP. For both datasets, we report Quadratic Weighted Kappa (QWK) scores, a widely adopted metric in AES  evaluation~\citep{ijcai2019p879}, as our measure of agreement between human scores and model outputs.
\paragraph{TOEFL11}
TOEFL11~\citep{blanchard2013toefl} contains 12,100 essays across eight essay prompts. Each essay is labeled with one of three proficiency levels (high, medium, low), which were converted from original 5-point scores. For our experiments, we used the official TOEFL 5-point scoring rubric to have LLMs generate 5-point scores, then followed~\citet{yoshida2025rubrics} to map these scores to three levels: scores 1-2 as "low," 3 as "medium," and 4-5 as "high." Following the official train/validation/test split, we sampled 100 essays each for training and validation sets, and used all 1,100 essays in the test set.

\begin{figure*}[t]
    \centering
    \begin{tcolorbox}[title=ASAP P1 Refined Rubric from Simplest Rubric (GPT-4.1),colframe = RoyalBlue, colback = RoyalBlue!10!White,]
... (omitted for brevity) ... \\
**General Guidelines \& Clarifications:**\\
- **Breadth and Depth for 4/5 Difference:**  \\
  - *Score 5* when the essay gives multiple (3+) reasons or arguments, and most are developed with more than a basic statement—some explanation, detail, or partial evidence is offered for each, even if language is weak and the support isn't fully detailed.\\
  ... (omitted for brevity) ... \\
  **Summary Table:**

| Score | Position      | Reasons/Development         | Support      | Organization   | Errors/Awkwardness   |
|-------|--------------|----------------------------|--------------|----------------|----------------------|
| 6     | Strong/clear | Fully developed, specific   | Thorough     | Excellent      | Very few             |
| 5     | Clear        |
\\... (omitted for brevity) ...
    \end{tcolorbox}
    \caption{ASAP P1 Refined Rubric from Simplest Rubric ``Based on the response’s content, rate the response on a scale of 1 to 6." (GPT-4.1)}
    \label{fig:asap_refined_rubric}
\end{figure*}

\paragraph{ASAP}
ASAP~\citep{ben2012asap} contains eight essay prompts, each with its own scoring rubric. We used Prompt 1 (P1), which is a 6-point scoring scale with a publicly available rubric. Following~\citet{lee-etal-2024-unleashing}, we sampled 10\% of the data (179 essays) as the test set, and sampled 100 essays each for training and validation from the remaining data. The ASAP dataset includes annotations from two human raters, and the "QWK between humans" shown in~\autoref{fig:experiments} represents the inter-annotator agreement.

We set the number of iterations \( T \) = 10 and batch size \( \mathcal{B} \) = 10. For each seed rubric and model, we ran three trials and save best rubric in validation set. We evaluated five language models: GPT-4.1~\citep{gpt4.1}, GPT-5-mini~\citep{gpt5}, Gemini-2.5-Flash, Gemini-2.5-Pro~\citep{gemini25}, and Qwen3-Next-80B-A3B-Instruct~\citep{qwen3-next}. We ran evaluation on test set three times and report the mean and standard deviation of QWK.~\autoref{app:setup} explains more details about experiments.

\section{Results}

\paragraph{Improvements from Rubric Refinement}
Across five models we tested, rubric refinement improved performance for four models on ASAP and for two models on TOEFL11 (see \autoref{fig:all_ets}, \ref{fig:all_asap} in \autoref{app:results}). As shown in \autoref{fig:experiments}, QWK increases by up to 0.47 in ASAP. Because TOEFL11 uses a coarse three-level scale (low/medium/high), its absolute QWK values are high overall. Whether the smaller pool of error cases available for refinement on TOEFL11, relative to ASAP, limits the attainable gains is an open question for future work. 

\paragraph{Refined Rubric}
As shown in ~\autoref{fig:asap_refined_rubric}, the refined rubrics add (i) visual emphasis (e.g., boldface) to highlight key evidence, (ii) a brief summary table at the end of the rubric, and (iii) explicit conditional rules of the form “if X is observed, assign score s.”

\paragraph{Seed Rubric}
As shown in \autoref{tab:seed_rubric}, even starting from a simplest rubric, ``Based on the response’s content, rate the response on a scale of 1 to 6.", the refined rubric can outperform a carefully crafted human-written rubric. This suggests that the model can infer discriminative scoring criteria from minimal guidance, potentially reducing the time and cost of authoring task-specific rubrics.

\section{Conclusion}
Inspired by prompt optimization and human rating process, we proposed an iterative method for AES where LLMs refine given scoring rubric to better align with human evaluation. Experiments on ASAP and TOEFL11 confirmed that this approach improves scoring agreement. Notably, even a rubric refined from a minimal starting point can outperform a detailed, human-authored one.

\section*{Limitations}
The experiments were conducted on two specific datasets, TOEFL11 and ASAP Prompt 1, so the findings may not generalize to other essay types or domains without further investigation. The refinement process needs 200 annotated samples for training and validation, which could be a practical barrier to broader adoption. The optimization was guided solely by maximizing the QWK metric, which may not capture all aspects of scoring quality and could limit gains on datasets with already high baseline scores.


\bibliography{custom}
\appendix
\section{Details of Experiments Setup}
\label{app:setup}
\autoref{fig:scoringprompt} shows scoring prompt in evaluation and \autoref{fig:refinementprompt} shows a refinement prompt for LLMs. \autoref{fig:exampleformat} shows the information for refinement. \autoref{fig:toefl_rubric_human}, \ref{fig:toefl_rubric_simplified}, \ref{fig:toefl_rubric_simplest} shows seed prompts for TOEFL 11. \autoref{fig:asap_rubric_human}, \ref{fig:asap_rubric_simplified}, \ref{fig:asap_rubric_simplest} shows seed prompts for ASAP P1. 

We set temperature = 1.0, max\_tokens = 8192 for GPT4.1, temperature = 1.0, reasoning max\_tokens = 0, max\_tokens = 8192 for Gemini-2.5-Flash, reasoning effort = "low", max\_tokens=8192 for GPT-5-mini, temperature = 1.0, reasoning max\_tokens = 1024, max\_tokens = 8192 for Gemini-2.5-pro, and temperature = 0.7, top\_p = 0.8, top\_k = 20 for Qwen3-Next-80B-A3B-Instruct.

GPT models were accessed via OpenAI's API service, while Gemini and Qwen3 models were accessed through VertexAI's API service routed by OpenRouter~\citep{openrouter}. Running experiments on Qwen3-Next-80B-A3B-Instruct for three seed rubrics and three trials of iterative refinements and evaluating three times on test set cost about \$25 (55 million tokens for input, 14 million tokens for completion).

\begin{figure}[t]
    \centering
    \begin{tcolorbox}[title=Scoring Prompt]
    You are a rater for writing responses on a high-stakes English language exam for second language learners. You will be provided with a prompt and the test-taker's response. Your rating should be based on the rubric below, following the specified format.

\# Essay Prompt\\
"""{essay\_prompt}"""\\
\# Response\\
"""{response}"""\\
\# Rubric\\
"""{rubric}"""\\
\# Output format:\\
Rationale: [<<<Your rationale here.>>>]\\
Rating: [<<<Your rating here.>>>]\\
    \end{tcolorbox}
    \caption{Scoring prompt.}
    \label{fig:scoringprompt}
\end{figure}

\begin{figure}[t]
    \centering
    \begin{tcolorbox}[title=Refinement Prompt]
I provided an assistant with the following rubrics to perform an essay grading task for me:\\
```
{current\_rubric}
```\\

The following are examples of different inputs to the assistant, the rationales for scores from the assistant, the scores from the assistant, and desired scores which I would like the assistant to achieve.\\
```
{examples}
```\\
Please analyze the rubrics and the examples, and then propose new rubrics that will help the assistant to perform better on this task.\\

Read all the assistant responses and reflect on the rationales given by the assistant. Identify any patterns or common themes in the rationales that led to correct or incorrect ratings. Consider how the rubrics could be adjusted to better align with these patterns by providing clearer/detailed guidelines for the assistant to follow.\\

Provide the new rubrics within ``` blocks.
    \end{tcolorbox}
    \caption{Refinement prompt.}
    \label{fig:refinementprompt}
\end{figure}

\begin{figure}[t]
    \centering
    \begin{tcolorbox}[title=Example Format]
Input for the assistant:\\
Essay Prompt:\\
"""{essay\_prompt}"""\\
Essay to be rated:\\
"""{response}"""\\
Rationale from the assistant:\\
"""{rationale}"""\\
Score from the assistant:\\
"""{rating}"""\\
Desired score:\\
"""{desired\_rating}"""\\
    \end{tcolorbox}
    \caption{Example format.}
    \label{fig:exampleformat}
\end{figure}

\begin{figure}[t]
    \centering
    \begin{tcolorbox}[title=TOEFL11 Rubric (Human),colframe = RoyalBlue, colback = RoyalBlue!10!White,]
\#\# Score 5\\
An essay at this level largely accomplishes all of the following:\\
- effectively addresses the topic and task\\
- is well organized and well developed, using clearly appropriate explanations, exemplifications, and/or details\\
- displays unity, progression, and coherence\\
- displays consistent facility in the use of language, demonstrating syntactic variety, appropriate word choice, and idiomaticity, though it may have minor lexical or grammatical errors\\
... (omitted for brevity) ... 
    \end{tcolorbox}
    \caption{TOEFL11 Rubric (Human)}
    \label{fig:toefl_rubric_human}
\end{figure}

\begin{figure}[t]
    \centering
    \begin{tcolorbox}[title=TOEFL11 Rubric (Simplified Human),colframe = RoyalBlue, colback = RoyalBlue!10!White,]
Score 5: Fully addresses the topic with clear organization, strong examples, smooth flow, and excellent language use.\\
Score 4: Addresses the topic well, with good organization and examples, though some points could be clearer.\\
... (omitted for brevity) ... 
    \end{tcolorbox}
    \caption{TOEFL11 Rubric (Simplified Human)}
    \label{fig:toefl_rubric_simplified}
\end{figure}

\begin{figure}[t]
    \centering
    \begin{tcolorbox}[title=TOEFL11 Rubric (Simplest),colframe = RoyalBlue, colback = RoyalBlue!10!White,]
Based on the response's content, rate the response on a scale of 1 to 5.
    \end{tcolorbox}
    \caption{TOEFL11 Rubric (Simplest)}
    \label{fig:toefl_rubric_simplest}
\end{figure}

\begin{figure}[t]
    \centering
    \begin{tcolorbox}[title=ASAP P1 Rubric (Human),colframe = RoyalBlue, colback = RoyalBlue!10!White,]
Score Point 1: An undeveloped response that may take a position but offers no more than very minimal support. Typical elements:\\
- Contains few or vague details.\\
- Is awkward and fragmented.\\
- May be difficult to read and understand.\\
- May show no awareness of audience.\\
... (omitted for brevity) ... \\
Note: \\
I have made an effort to remove personally identifying information from the essays using the Named Entity Recognizer (NER). The relevant entities are identified in the text and then replaced with a string such as "PERSON", "ORGANIZATION", "LOCATION", "DATE", "TIME", "MONEY", "PERCENT", "CAPS" (any capitalized word) and "NUM" (any digits). Please do not penalize the essay because of the anonymizations.
    \end{tcolorbox}
    \caption{ASAP P1 Rubric (Human)}
    \label{fig:asap_rubric_human}
\end{figure}

\begin{figure}[t]
    \centering
    \begin{tcolorbox}[title=ASAP P1 Rubric (Simplified Human),colframe = RoyalBlue, colback = RoyalBlue!10!White,]
Score Point 6: A well-developed response that takes a clear and thoughtful position and provides persuasive support.\\
... (omitted for brevity) ...
    \end{tcolorbox}
    \caption{ASAP P1 Rubric (Simplified Human)}
    \label{fig:asap_rubric_simplified}
\end{figure}

\begin{figure}[t]
    \centering
    \begin{tcolorbox}[title=ASAP P1 Rubric (Simplest),colframe = RoyalBlue, colback = RoyalBlue!10!White,]
Based on the response's content, rate the response on a scale of 1 to 6.
    \end{tcolorbox}
    \caption{ASAP P1 Rubric (Simplest)}
    \label{fig:asap_rubric_simplest}
\end{figure}

\section{Detailed Results}
\label{app:results}
\begin{figure*}[t]
\centering
\includegraphics[width=\linewidth]{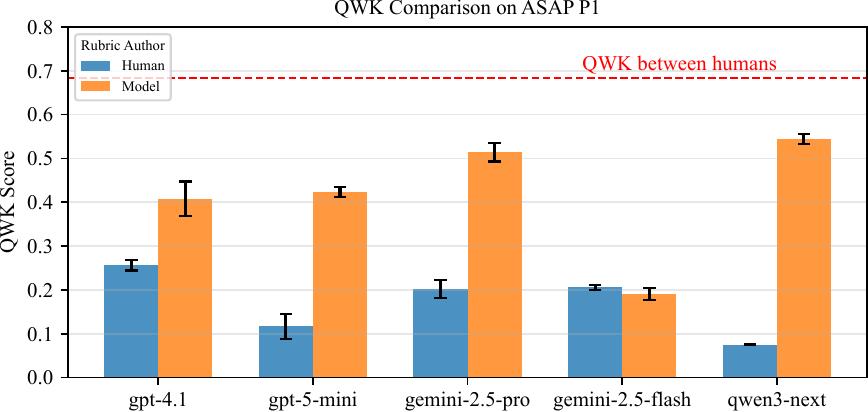}
  \caption{QWK comparison using refined rubric best scored in validation set and using human written rubric(\texttt{human\_rubric}) on ASAP P1.}
  \label{fig:all_asap}
\end{figure*}

\begin{figure*}[t]
\centering
\includegraphics[width=\linewidth]{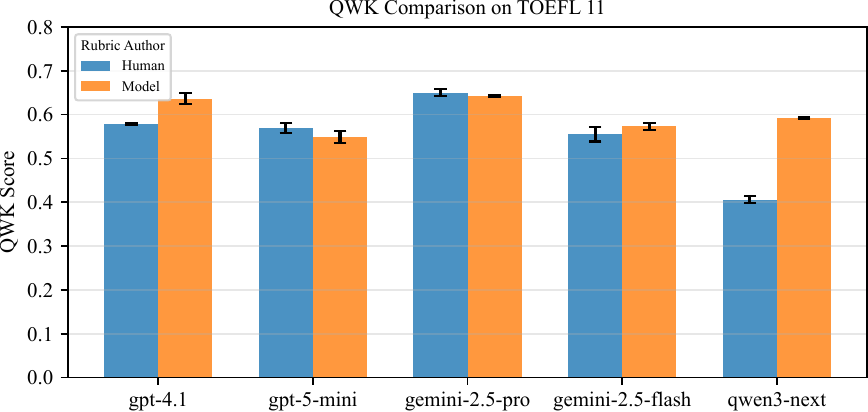}
  \caption{QWK comparison using refined rubric best scored in validation set and using human written rubric(\texttt{human\_rubric}) on TOEFL11.}
  \label{fig:all_ets}
\end{figure*}

\autoref{fig:all_asap}, \ref{fig:all_ets} shows QWK comparison of all models using refined rubric best scored in validation set and using human written rubric(\texttt{human\_rubric}) on ASAP P1 and TOEFL11.


\end{document}